\documentclass[10pt,twocolumn,letterpaper]{article}

\usepackage{cvpr}              

\renewcommand{\paragraph}[1]{\vspace{.4em}\noindent\textbf{#1.}}

\usepackage{xspace}
\newcommand{\ours}{\textsc{HierAmp}\xspace}

\usepackage{multirow}
\usepackage{array} 
\usepackage{booktabs} 
\usepackage{makecell} 
\usepackage[table]{xcolor}
\usepackage{algorithm}
\usepackage{algpseudocode}
\usepackage{fontawesome}

\definecolor{cvprblue}{rgb}{0.21,0.49,0.74}
\usepackage[pagebackref,breaklinks,colorlinks,allcolors=cvprblue]{hyperref}

\def\paperID{2478} 
\def\confName{CVPR}
\def\confYear{2026}

\title{\ours: Coarse-to-Fine Autoregressive Amplification\\ for Generative Dataset Distillation} 

\author{
\begin{tabular}{c}
Lin Zhao$^{1*}$ \quad
Xinru Jiang$^{1*}$ \quad
Xi Xiao$^{2}$ \quad
Qihui Fan$^{1}$ \quad
Lei Lu$^{1}$ \\
Yanzhi Wang$^{1}$ \quad
Xue Lin$^{1}$ \quad
Octavia Camps$^{1}$ \quad
Pu Zhao$^{1\dagger}$ \quad
Jianyang Gu$^{3\dagger}$ 
\end{tabular}
\\[6pt]
{\normalsize$^1$ Northeastern University \quad $^2$ University of Alabama at Birmingham \quad $^3$ The Ohio State University}\\
}

\begin{document}

\twocolumn[{
\renewcommand\twocolumn[1][]{#1}
\maketitle
\begin{center}
    \centering
    \captionsetup{type=figure}
    \includegraphics[width=0.9\linewidth]{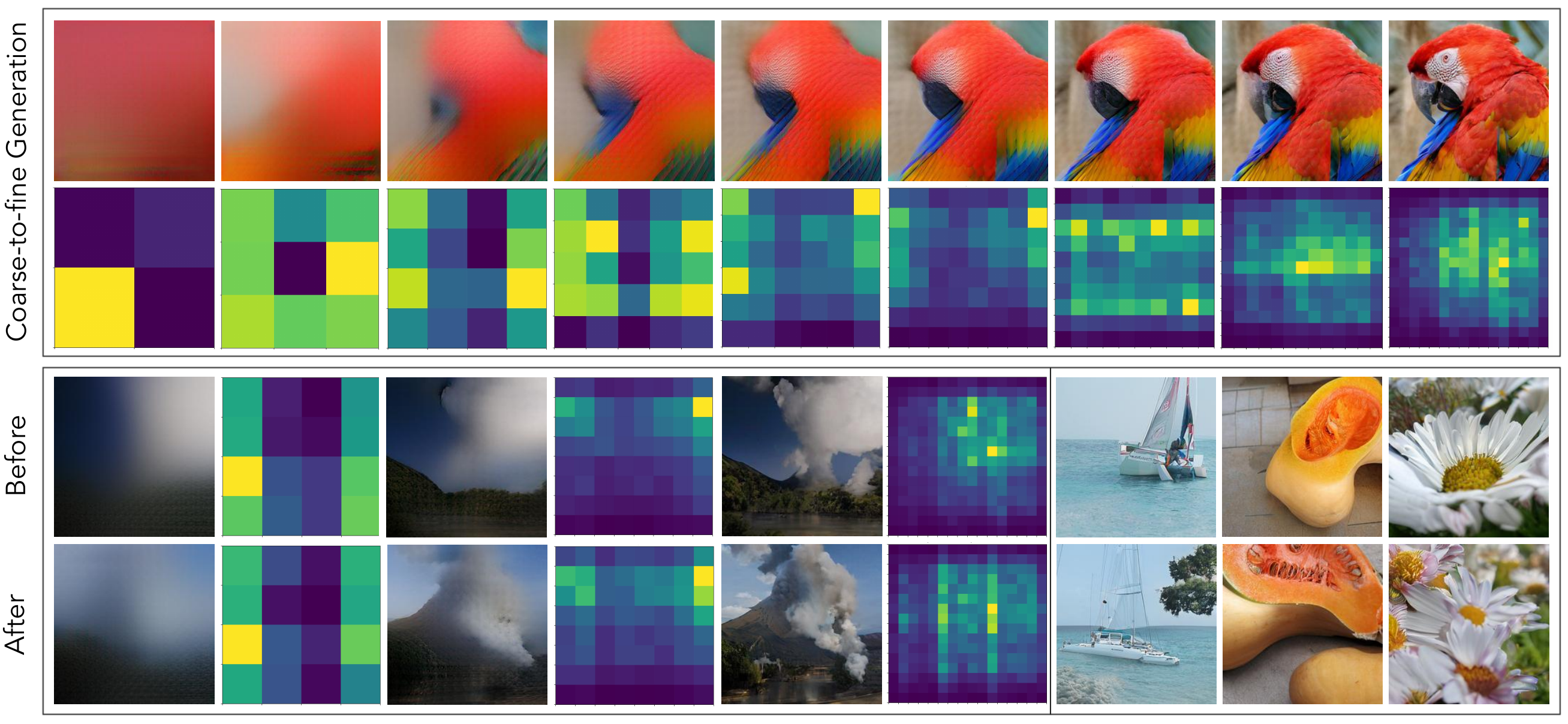}
    \caption{\textbf{Top}: The visual autoregressive model constructs coarse scene structure and gradually complements details from coarse to fine scales. The class token highlights regions that express object-related semantics (the second row). \textbf{Bottom}: \ours identifies important semantic regions and refines the hierarchical structure and details in an autoregressive manner. The images after amplification demonstrate more diverse components and richer class-related details.}
    \label{fig: teasor}
\end{center}
}]

\begingroup
\renewcommand\thefootnote{}
\footnote{$^{*}$Equal contribution; $^\dagger$ Corresponding author}
\addtocounter{footnote}{-1}
\endgroup

\begin{abstract}
Dataset distillation often prioritizes global semantic proximity when creating small surrogate datasets for original large-scale ones. However, object semantics are inherently hierarchical. For example, the position and appearance of a bird's eyes are constrained by the outline of its head. Global proximity alone fails to capture how object-relevant structures at different levels support recognition. In this work, we investigate the contributions of hierarchical semantics to effective distilled data. We leverage the vision autoregressive (VAR) model whose coarse-to-fine generation mirrors this hierarchy and propose \ours to amplify semantics at different levels. 
At each VAR scale, we inject class tokens that dynamically identify salient regions and use their induced maps to guide amplification at that scale. This adds only marginal inference cost while steering synthesis toward discriminative parts and structures.
Empirically, we find that semantic amplification leads to more diverse token choices in constructing coarse-scale object layouts. Conversely, at fine scales, the amplification concentrates token usage, increasing focus on object-related details. Across popular dataset distillation benchmarks, \ours consistently improves validation performance without explicitly optimizing global proximity, demonstrating the importance of semantic amplification for effective dataset distillation. 
\faGithub\ \url{https://github.com/Oshikaka/HIERAMP}


%
%
%
%

\end{abstract}    
\section{Introduction}
\label{sec:intro}

Dataset distillation aims to synthesize a small surrogate dataset from a large training corpus while preserving downstream performance. Most previous efforts optimize global distributional proximity, where features or training dynamics between synthetic and real data are matched \cite{cazenavette2022dataset, cui2023scaling, luo2024llm, du2024sequential}.
While reproducing the overall distribution, the distillation process does not directly reflect key factors that influence downstream performance. A distilled set may look close to the original set, but underrepresent the discriminative semantics that models use to separate classes. 

In this work, we investigate dataset distillation from the perspective of object semantics, as a complement to distributional proximity. 
The semantics of a specific object in an image are inherently hierarchical \cite{Peng_2023_CVPR, ma2022open, Zhong_2025_CVPR}. 
As shown in \ref{fig: teasor}, the global layout governs the coarse-level spatial organization and object placement. At a finer granularity, the semantics of individual parts constrain the associated textures and details.
Vision autoregressive (VAR) models naturally reflect this characteristic by synthesizing images in a coarse-to-fine manner \cite{tian2024visual, ren2024m}. Early scales generate the overall structure, while deeper scales focus more on subtle details. Our method exploits this alignment and explores the effect of object-related semantics at different scales. 

Based on this, we propose \ours to amplify the generation process in an autoregressive fashion. Concretely, we inject learnable class tokens into each scale of the VAR model and optimize them with a classification objective \cite{zhang2022dino}. 
During generation, the class token of scale $\ell$ aggregates context and produces a soft importance map over spatial tokens. 
The map highlights regions with object-related semantics expressed at that scale. 
Therefore, we amplify attention toward tokens with higher importance scores during autoregressive decoding.
Compared with adopting external segmentation tools \cite{liu2024grounding}, the design adds only a marginal inference cost and avoids heavy guidance at test time. More importantly, it allows for more fine-grained salient identification across different generation scales. 
This framework enables the analysis of semantics at different hierarchical levels. We examine the change of token-distribution diagnostics (\eg, entropy and token coverage) before and after amplifying the attention along the VAR scales. Empirically, we find that the amplification leads to distinct effects at different generation scales. For coarse scales, the token distribution becomes more uniform and diverse. Oppositely, the amplification at fine scales concentrates the token usage. 
By comparing the validation performance, we find that amplifying coarse scales leads to the most significant accuracy improvement. Qualitatively, we show that while the coarse scales do not directly contribute to object-specific details, they set the overall structure and largely influence the semantic richness of later scales. 

Extensive experiments are conducted across popular dataset distillation benchmarks, where \ours achieves state-of-the-art validation accuracy. \ours uncovers the relationship between hierarchical semantics and downstream model training, which enhances the explainability of dataset distillation. Through this work, we call for more attention toward understanding the underlying mechanisms that support effective and trustworthy dataset distillation. 




\section{Related Works}
\label{sec:related works}

\subsection{Dataset Distillation}
Dataset Distillation (DD) \cite{wang2018dataset} compresses a large training set into a small synthetic set that preserves training performance. Prior works formulate dataset distillation as a bi-level optimization problem \cite{zhao2020dataset, cazenavette2022dataset}, mainly via gradient matching \cite{zhao2020dataset, zhao2021dataset, lee2022dataset, kim2022dataset, vahidian2025group} or trajectory matching \cite{cazenavette2022dataset, cui2023scaling, du2023minimizing,du2024sequential}. 
However, these methods are computationally expensive and  difficult to scale to high-resolution images and large datasets, with limited cross-architecture generalization. 
To address this, distribution matching \cite{zhao2023dataset,zhao2023improved,deng2024exploiting} aligns feature statistics in embedding spaces, shifting from explicit optimization to statistical alignment. Building on this, recent efficient distillation methods \cite{cui2023scaling, shao2024generalized, sun2024diversity, kim2022dataset} further improve scalability and performance. 
However, images produced by these approaches often lack visual fidelity and appear perceptually unrealistic, resembling feature abstractions rather than natural images.

The above limitation motivates the adoption of generative models that prioritize visual realism and fidelity beyond purely optimization-driven objectives.
Early GAN-based methods \cite{wang2024dim, sajedi2024data, Zhong_2025_CVPR} produce more representative samples. However, they exhibit limited data diversity, which hinders effective distillation. More recently, Diffusion models \cite{gu2024efficient, Su_2024_CVPR, wang2025cao2, zhao2025taming, gu2025concord} have emerged as the state-of-the-art approach, providing more higher quality, diverse samples that better preserve the characteristics of the original dataset. 

\begin{figure*}[t]
  \centering
  \includegraphics[width=0.85\textwidth]{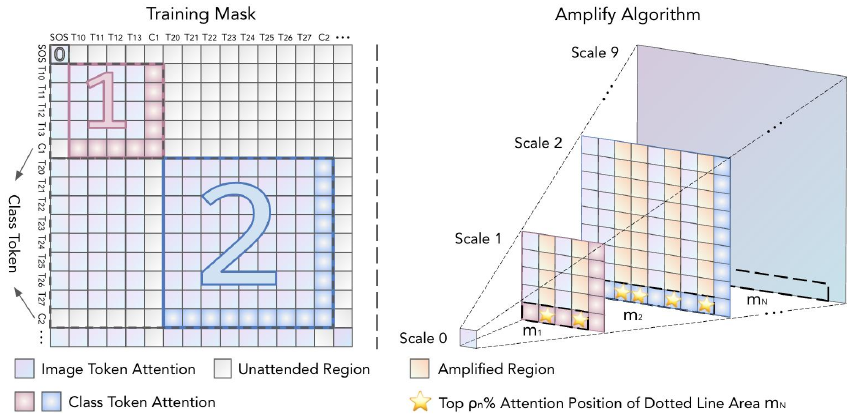}
  \caption{\textbf{Overview of the \ours framework.} 
  \textbf{Left:} Scale-Restricted Class Token Attention Mask. The class token attends only to image tokens from the corresponding scale, with grey regions indicating blocked attention, producing a scale-specific semantic summary. 
  \textbf{Right:} Multi-Scale Semantic Feature Amplification. The Amplify Algorithm selects the top attention positions from the class-token map at each scale and amplifies them via a positive logit bias, guiding the model to focus on semantically important features during decoding.}

  \label{fig:method}
\end{figure*}

\subsection{Generative Model}
Generative modeling aims to learn data distributions for realistic synthesis \cite{goodfellow2016deep}. GANs \cite{goodfellow2014generative} achieve high fidelity but often suffer from mode collapse and unstable training at high resolutions \cite{adler2018banach}. To improve stability and sample diversity, diffusion models \cite{ho2020denoising,songdenoising} are introduced, producing high-quality and diverse results across domains \cite{dhariwal2021diffusion,zhao2026s2dit,zhao2024flasheval}. However, their long denoising chains incur substantial computational cost \cite{croitoru2023diffusion}.
In contrast, Visual Autoregressive (VAR) models \cite{tian2024visual} introduce a coarse-to-fine hierarchy for scale-aware control, achieving competitive quality with fewer sampling steps and providing a strong generative backbone.

\section{Method}

\subsection{Preliminary} \label{sec:pre}

Rather than predicting the next token as in other autoregressive generators~\cite{li2024autoregressive,sun2024autoregressive,ControlAR}, 
VAR~\cite{tian2024visual} predicts the next scale over the entire token map.
Since the model operates in a discrete token space, VAR employs a VQ-VAE~\cite{van2017neural} to bridge token features and images via codebook dequantization and decoding.
When predicting the feature of current scale $r_n$, VAR uses the ground-truth feature $\big(r_1,\ldots, r_{n-1}\big)$.
During training, it minimizes the cross-entropy loss over all scales with teacher forcing as  follows:
\begin{equation}
P \;=\; \prod_{n=1}^{N} p_{\theta}\!\left(r_n \mid r_1,\ldots, r_{n-1}\right),
\end{equation}
where $P$ is the predicted feature of all scales, and $p_{\theta}$ indicates the VAR model.
For the attention in transformer blocks, it adopts a scale-based masking to ensure that the tokens at scale $n$ can only attend to earlier scales.

During inference, the model samples one scale at a time. The $N$ scale generation can be expressed as:
\begin{equation}
\begin{aligned}
r_1 &\sim p_{\theta}\!\left(r_1 \mid s\right),\\
r_2 &\sim p_{\theta}\!\left(r_2 \mid r_{1}\right),\\ &\ldots\\
r_N &\sim p_{\theta}\!\left(r_N \mid r_1,\ldots, r_{N-1}\right),
\end{aligned}
\end{equation}
where $s$ is the initial class embedding.
Notably, after generating the $n^{th}$ scale, $r_n$ is incorporated via residual addition with the upsampling feature of the $(n-1)^{th}$ scale, yielding the updating for the  $(n+1)^{th}$ scale:
\begin{equation}
\begin{aligned}
r_n &= r_n + \mathcal{U}_{(n-1)\to n}\!\big(r_{n-1}\big),\\
\end{aligned}
\end{equation}
where $\mathcal{U}_{a\to b}(\cdot)$ denotes the upsampling operator that maps features from scale $a$ to scale $b$.

\noindent\textbf{Intuition.}
VAR learns ``what remains'' at each finer scale and composes an image by successively adding those coarse-level layout first, then mid-level structure, and finally fine-level details.

\subsection{Motivation}
Current dataset distillation methods mainly operate on images monolithically, which directly match the original data distribution in pixel space \cite{cazenavette2022dataset} or latent space \cite{zhao2025taming, Su_2024_CVPR}. 
There are two main drawbacks for these methods: (i) The feature mapping between surrogate dataset and full dataset is typically performed in a low-level structural feature space, with limited semantic understanding. (ii) All features are modeled in a single latent space, without accounting for the hierarchical nature of image information \cite{Zhong_2025_CVPR, li2025hyperbolic}.
This motivates us to analyze the problem from a coarse-to-fine semantic generation perspective, to identify hierarchical designs that better benefit the dataset distillation task.
Consequently, the coarse-to-fine nature of VAR is tightly aligned with our objective of modeling semantics across scales.

\subsection{Semantic-guided Attention Analysis}
VAR is a generative model based on transformer blocks, in which self-attention provides a natural probe into how information flows during generation. 
Therefore, we analyze the attention features from a semantic perspective to understand what each scale encodes and how to optimize the synthesis for dataset distillation.

We begin by extracting semantic patterns from attention maps as the basis for our analysis.
To achieve this, we introduce learnable class tokens following DINO~\cite{caron2021emerging,oquab2023dinov2}.
As VAR performs residual refinement across multiple scales, the regions of focus vary by scale. Accordingly, we introduce a learnable class token at each scale to capture the semantic information.
As shown in \cref{fig:method}-left, at scale $n$, the class token is constrained by a scale-restricted attention mask only attend to tokens from the same scale, yielding a scale-specific semantic summary.
Notably, while VAR allows regular tokens at scale $n$ to attend to tokens from earlier scales ($1\!:\!n\!-\!1$), the class token is explicitly masked to ignore such cross-scale connections and focus exclusively on the current scale.

%
The attention map at scale $n$ is denoted by $\mathbf{X}_n\in\mathbb{R}^{L_{q}^n\times L_{k}^n}$,  and $[c]_n\in\mathbb{R}^{d}$ denotes a learnable class token.
We append $[c]_n$ to obtain
$\tilde{\mathbf{X}}_n=[\,\mathbf{X}_n , [c]_n\,]$. For multi-head attention with
$H$ heads, the query $\mathbf{Q}^{(h)}_{n}$, key $\mathbf{K}^{(h)}_{n}$ and value $\mathbf{V}^{(h)}_{n}$ for head $h$ at scale $n$ are computed as:
\begin{equation}
\scalebox{0.9}{$
\mathbf{Q}^{(h)}_{n}=\tilde{\mathbf{X}}_{n}\mathbf{W}^{(n)}_{Q},\quad
\mathbf{K}^{(h)}_{n}=\tilde{\mathbf{X}}_{n}\mathbf{W}^{(n)}_{K},\quad
\mathbf{V}^{(h)}_{n}=\tilde{\mathbf{X}}_{n}\mathbf{W}^{(n)}_{V},
$}
\end{equation}
where $\mathbf{W}^{(n)}_{Q}$, $\mathbf{W}^{(n)}_{K}$ and $\mathbf{W}^{(n)}_{V}$ are learnable projection matrices. Based on this, the attention map between class-token query and same-scale keys for head $h$ can be  expressed as:
\begin{equation}
\scalebox{0.98}{$
\boldsymbol{\alpha}^{(h)}_{n,\mathrm{cls}}
=\operatorname{softmax}\!\left(
\frac{\mathbf{Q}^{(h)}_{n}[:, -1]\;(\mathbf{K}^{(h)}_{n})^{\top}}{\sqrt{d_h}}
+\mathbf{m}^{(h)}_{n,\mathrm{cls}}
\right),
$}
\end{equation}
where $\mathbf{m}^{(h)}_{n,\mathrm{cls}}\in\{0,-\infty\}^{1\times(1+L_k)}$ is zero on positions from scale $n$ and $-\infty$ on all other positions.
Given the class-token attention map $\boldsymbol{\alpha}^{(h)}_{n,\mathrm{cls}}$, we form the class token embedding $\mathbf{c}^{e}_{n}$, and apply a lightweight classifier $p_n(.)$ to train the $[c]_n$:
\begin{equation}
\mathcal{L}_{\mathrm{cls}}
= \frac{1}{N}\sum_{n=1}^{N}\Big(-\log p_{n}(\mathbf{c}^{e}_{n})\Big).
\end{equation}
By assigning a unique class token to each scale, we explicitly learn the semantic information at every scale.
We use  $\boldsymbol{\alpha}^{(h)}_{n,\mathrm{cls}}$ as a semantic saliency map for scale $n$,
aggregated over heads:
\begin{equation}
\begin{aligned}
\mathbf{m}_n &\,=\, \frac{1}{H}\sum_{h=1}^{H}\boldsymbol{\alpha}^{(h)}_{n,\mathrm{cls}}
\;\in\; \mathbb{R}^{1\times L_k},\\
\mathbf{M}_n &\,=\, \operatorname{R}_{h_n\times w_n}(\mathbf{m}_n)
\;\in\; \mathbb{Re}^{h_n\times w_n},
\end{aligned}
\end{equation}
where $\mathbf{m}_n\!\in\!\mathbb{R}^{1\times L_k}$ denotes the class-token attention map aggregated over heads. $\operatorname{Re}$ means that reshape $\mathbf{m}_n$ to $\mathbf{M}_n\!\in\!\mathbb{R}^{h_n\times w_n}$, which aligns with the $h_n\times w_n$ token grid of the image height and weight at scale $n$, with $h_n w_n = L_k$.
%
 \cref{fig: teasor} visualizes \(\mathbf{M}_n\) at different scales.
The highlighted areas indicate that  $[c]_n$ successfully captures semantically important regions.



\subsection{Coarse-to-fine Autoregressive Amplification}

\begin{table*}[ht]
\centering
\resizebox{\textwidth}{!}{
\begin{tabular}{lc|ccccc|ccccc}
\toprule
\multirow{2}{*}{Dataset}& \multirow{2}{*}{IPC} & \multicolumn{5}{c}{ResNet-18}
& \multicolumn{5}{|c}{ResNet-101}  \\
\cmidrule(lr){3-7} \cmidrule(lr){8-12}
 & & Minimax & D$^3$HR & RDED & CaO$_2$ & Ours 
   & Minimax & D$^3$HR & RDED & CaO$_2$ & Ours\\
\midrule
\multirow{2}{*}{CIFAR-10} 
& 10 & $-$ & $41.3\pm0.1$ & $37.1 \pm 0.3$ & $39.0 \pm 1.5$ & $\mathbf{44.3}\pm\mathbf{0.6}$
      & $-$ & $35.8\pm0.6$ & $33.7\pm0.3$ & $-$ & $\mathbf{64.3}\pm\mathbf{0.9}$\\
& 50 & $-$ & $70.8\pm0.5$ & $ 62.1\pm 0.1$ & $-$ & $ \mathbf{72.0} \pm \mathbf{0.1} $
      & $-$ & $63.9\pm0.4$ & $51.6\pm0.4$ & $-$ & $\mathbf{64.3}\pm \mathbf{0.9}$\\
\midrule
\multirow{2}{*}{CIFAR-100} 
& 10 & $-$ & $49.4\pm0.2$ & $42.6\pm 0.2$ & $-$ & $\mathbf{52.0}\pm\mathbf{0.1}$
      & $-$ & $46.0\pm0.5$ & $41.1\pm0.2$ & $-$ & $\mathbf{49.7}\pm\mathbf{0.2}$\\
& 50 & $-$ & $65.7\pm0.3$ & $62.6\pm 0.1$ & $-$ & $ \mathbf{66.5} \pm \mathbf{0.2} $
      & $-$ & $\mathbf{66.6}\pm\mathbf{0.2}$ & $63.4\pm0.3$ & $-$ & $66.1\pm 0.2$\\
\midrule
\multirow{3}{*}{ImageNet-Woof} 
& 1  & $19.9\pm0.2$ & $-$ & $20.8\pm1.2$ & $\mathbf{21.1}\pm\mathbf{0.6}$ & $20.9\pm0.8$
      & $17.7\pm0.9$ & $-$ & $19.6\pm1.8$ & $21.2\pm1.7$ & $20.3\pm1.4$ \\
& 10 & $40.1\pm1.0$ & $39.6\pm1.0$ & $38.5 \pm 2.1$ & $45.6  \pm 1.4$ & $\mathbf{45.8}\pm\mathbf{1.6}$
      & $34.2\pm1.7$ & $-$ & $31.3\pm1.3$ & $36.5\pm1.4$ & $\mathbf{39.0}\pm\mathbf{1.4}$\\
& 50 & $67.0\pm1.8$ & $57.6\pm0.4$ & $68.5 \pm 0.7$ & $68.9 \pm 1.1$ & $ \mathbf{70.0} \pm \mathbf{0.8} $
      & $62.7\pm1.6$ & $-$ & $59.1\pm0.7$ & $63.1\pm1.3$ & $\mathbf{66.2}\pm \mathbf{1.2}$\\
\midrule
\multirow{3}{*}{ImageNet-100} 
& 1  & $7.3\pm0.1$ & $-$ & $8.1\pm0.3$ & $8.8\pm0.4$ & $\mathbf{8.9}\pm\mathbf{0.3}$
      & $5.4\pm0.6$ & $-$ & $6.1\pm0.8$ & $\mathbf{6.6}\pm\mathbf{0.4}$ & $6.24\pm 0.4$\\
& 10 & $32.0\pm1.0$ & $-$ & $36.0\pm0.3$ & $36.6 \pm 0.2$ & $\mathbf{36.7}\pm\mathbf{0.3}$
      & $29.2\pm1.0$ & $-$ & $33.9\pm0.1$ & $34.5\pm0.4$ & $ \mathbf{35.1}\pm \mathbf{0.2} $\\
& 50 & $63.9\pm0.1$ & $-$ & $61.6\pm0.1$ & $68.0\pm0.5$ & $ \mathbf{68.1}\pm \mathbf{0.2}$
      & $67.4\pm0.6$ & $-$ & $66.0\pm0.6$ & $\mathbf{70.8}\pm\mathbf{0.2}$ & $ 66.1 \pm 0.4 $\\
\midrule
\multirow{4}{*}{ImageNet-1K} 
& 1   & $5.9\pm0.2$ & $-$ & $6.6\pm0.2$ & $7.1\pm0.1$ & $\mathbf{7.7}\pm\mathbf{0.1}$
       & $4.0\pm0.5$ & $-$ & $5.9\pm0.4$ & $6.0\pm0.4$ & $\mathbf{6.8}\pm\mathbf{0.4}$\\
& 10  & $44.3\pm0.5$ & $44.3\pm0.3$ & $42.0\pm0.1$ & $46.1\pm0.2$ & $\mathbf{47.6}\pm\mathbf{0.1}$
       & $46.9\pm1.3$ & $\mathbf{52.1}\pm\mathbf{0.4}$ & $48.3\pm1.0$ & $52.2\pm1.1$ & $50.8\pm 0.4$\\
& 50  & $58.6\pm0.3$ & $59.4\pm0.1$ & $56.5\pm0.1$ & $60.0\pm0.0$ & $\mathbf{60.8}\pm\mathbf{0.5}$
       & $65.5\pm0.1$ & $66.1\pm0.1$ & $61.2\pm0.4$ & $66.2 \pm 0.1$ & $\mathbf{66.4}\pm\mathbf{0.3}$\\
& 100 & $-$ & $62.5 \pm 0.0$ & $-$ & $-$ & $\mathbf{62.7}\pm\mathbf{0.3}$
       & $-$ & $68.1\pm0.0$ & $-$ & $-$ & $\mathbf{68.5}\pm \mathbf{0.3}$\\
\bottomrule
\end{tabular}
}
\caption{\textbf{Top-1 accuracy comparison with four SOTA methods}. As in RDED \cite{sun2024diversity}, all methods adopt ResNet-18 as the teacher and are trained on both ResNet-18 and ResNet-101. ‘–’ denotes missing results in the original paper.}
\label{tab:performance_comparison}
\end{table*}




As shown in \cref{fig:method}-right, we leverage the semantic class-token attention to guide autoregressive decoding from coarse to fine scales. 
Our model contains ten hierarchical scales indexed by $n\!\in\!\{0,\dots,9\}$. 
Since scale $0$ contains only a single patch (one token), we exclude it from amplification and operate on scales $n\!\in\!\{1,\dots,9\}$. We group this into three stages: \textit{\textbf{Coarse}} ($1$–$3$), \textit{\textbf{Mid}} ($4$–$6$), and \textit{\textbf{Fine}} ($7$–$9$), with \textit{\textbf{Full}} referring to all scales 1–9.
Since the high-magnitude entries of $\mathbf{m}_n$ correspond to locations most attended by the class token $[c]_n$, we select a set of salient positions $\mathcal{S}_n$ from $\mathbf{m}_n$ by keeping the top $\rho_n\%$ entries:
\begin{equation}
\mathcal{S}_n \;=\; \operatorname{Top}\text{-}\rho_n(\mathbf{m}_n).
\label{eq:cfa-saliency}
\end{equation}
Thus  a binary indicator $\mathbf{a}_n \in \{0,1\}^{L_k^n}$ on $\mathbf{m}_n$ can be obtained  according to $\mathcal{S}_n$:
\begin{equation}
(\mathbf{a}_n)_j =
\begin{cases}
1, & \text{if}\ \ j\in\mathcal{S}_n,\\
0, & \text{otherwise},
\end{cases}
\qquad j=1,\dots,L_k^n.
\end{equation}
Our goal is to steer attention toward semantically relevant regions—namely the positions $j$ with $(\mathbf{a}_n)_j=1$.

To make all queries at scale $n$ preferentially attend to these salient keys, we add positive logit bias to the corresponding key columns for every head $h$:
\begin{equation}
\label{eq:cfa-bias}
\begin{aligned}
\mathbf{L}^{(h)}_{n} &= \frac{\mathbf{Q}^{(h)}_{n}(\mathbf{K}^{(h)}_{n})^{\top}}{\sqrt{d_h}} + \mathbf{p}^{(h)}_{n},\\
\tilde{\mathbf{L}}^{(h)}_{n} &= \mathbf{L}^{(h)}_{n} + \beta_n\, \mathbf{1}_{L_k^n+1}\, \mathbf{a}_n^{\top}.
\end{aligned}
\end{equation}
where $\mathbf{p}^{(h)}_{n}$ denotes the original mask, $\mathbf{1}_{L_k^n+1}$ is an all-ones column vector matching the number of queries at the scale (including the class token), and $\beta_n\!>\!0$ controls the amplification strength.
The modified attention then becomes $\tilde{\boldsymbol{\alpha}}^{(h)}_{n}\!=\!\operatorname{softmax}(\tilde{\mathbf{L}}^{(h)}_{n})$, which increases the probability mass on semantically meaningful regions for every token at scale $n$.
We apply this procedure from coarse to fine, using a stage-aware schedule
\(
\rho_{1:3}, \rho_{4:6}, \rho_{7:9},\ 
\)
so early scales emphasize global object regions while later scales refine fine textures. 
This coarse-to-fine reinforcement aligns the attention hierarchy with semantic structure and improves semantic consistency and details across generation pipeline.
\section{Experiments}
\subsection{Experimental Settings}
\noindent \textbf{Datasets}.
We evaluate the practical applicability of our method on both large-scale and small-scale datasets. Our primary benchmark is ImageNet-1K ($224\times224$)~\cite{deng2009imagenet}, which contains 1,000 classes and approximately one million images. To assess performance under limited data conditions, we also use CIFAR-10 and CIFAR-100 ($32\times32$)~\cite{krizhevsky2009learning}, ImageWoof \cite{Howard_Imagewoof_2019}, a subset of 10 dog breeds, and ImageNet-100 \cite{russakovsky2015imagenet}, which includes 100 randomly selected classes from ImageNet-1K. To explore higher compression ratios while maintaining generalizability, we experiment with three images-per-class (IPC) settings: 1, 10, and 50, across all datasets, and  include an IPC of 100 for ImageNet-1K.

\noindent \textbf{Network Architectures}.
Following previous works \cite{sun2024diversity, zhao2025taming}, we evaluate our method on a variety of neural network architectures to confirm its performance. The experiments include ResNet-18 \& ResNet-101 \cite{he2016deep}, MobileNet-V2 \cite{sandler2018mobilenetv2}, and EfficientNet-B0 \cite{tan2019efficientnet}. All experiments are conducted three times to ensure fair and reliable comparisons with other methods.

\noindent \textbf{Baselines.}
We compare \ours against four state-of-the-art methods: Minimax \cite{gu2024efficient}, which formulates distillation as a minimax optimization to enhance generalization by utilizing diffusion models; D$^3$HR \cite{zhao2025taming}, a DDIM inversion approach that improves high-resolution synthesis by distribution matching; RDED \cite{sun2024diversity}, which preserves visual realism by selecting and cropping informative patches directly from real images; and CaO$_2$ \cite{wang2025cao2}, a diffusion-driven method that combines probabilistic sample selection with latent-code refinement to enhance conditional likelihood.

\noindent \textbf{Implementation Details.}
We adopt the pre-trained Visual Autoregressive Model (VAR) \cite{tian2024visual} with a depth of 16 and the resolution of $256\times256$, originally trained on ImageNet. The model is then fine-tuned for 5 epochs with a class token for semantic-guided attention. All experiments are conducted on NVIDIA RTX A6000 GPUs.

\subsection{Comparison with State-of-the-art Methods}
We evaluate the effectiveness of \ours against four state-of-the-art dataset distillation approaches: Minimax \cite{gu2024efficient}, D$^3$HR \cite{zhao2025taming}, RDED \cite{sun2024diversity}, and CaO$_2$ \cite{wang2025cao2}. Following the RDED evaluation protocol, all methods use ResNet-18 as the teacher and are separately trained and evaluated on ResNet-18 and ResNet-101. Missing results (‘–’) indicate entries not reported in the original papers. As shown in Table \ref{tab:performance_comparison}, \ours  achieves the highest performance across most datasets and IPC settings.

\noindent \textbf{Small-scale Datasets.} 
For CIFAR-10, our method reaches 44.3\% and 72.0\% on ResNet-18 at IPC$=10$ and $50$, respectively, outperforming the previous best method D$^3$HR. 
On CIFAR-100, \ours also achieves the highest accuracy at IPC$=10$ and remains competitive with the strongest baselines at IPC$=50$. On ImageNet-Woof, for IPC$=$10, it outperforms RDED by 6.2\% on ResNet-18 and 7.7\% on ResNet-101. Our method surpasses all other baselines at both IPC$=$10 and IPC$=$50, reaching 70.0\% at IPC$=$50.

\noindent \textbf{Mid-scale Datasets.} When scaling to ImageNet-100, it outperforms RDED and CaO$_2$ across most IPCs, achieving the highest accuracy of 68.14\% on ResNet-18 at IPC=50.

\noindent \textbf{Large-scale Datasets.} On the large-scale ImageNet-1K, it demonstrates a clear advantage. At IPC=1, it exceeds Minimax by 1.8\% and RDED by 1.1\% on ResNet-18. 
At IPC=10, it achieves 47.6\% on ResNet-18, outperforming the second-best method, CaO$_2$, by 1.5\%, while maintaining competitive performance on ResNet-101. At high IPCs (50 and 100), \ours consistently  surpasses all baselines, highlighting its robust scalability to larger datasets.

Besides, we report Frechet Inception Distance (FID) \cite{heusel2018ganstrainedtimescaleupdate} and computational efficiency comparisons with other methods, which can be seen in Appendix~\ref{sec:fid} and \ref{sec:latency}.

\subsection{Cross-architecture Generalization}
\begin{table}[t]
\centering
\footnotesize
\renewcommand{\arraystretch}{1.0}
\resizebox{0.5\textwidth}{!}{
\begin{tabular}{lc|ccc}
\toprule
\multicolumn{2}{c|}{Student\textbackslash Teacher} & ResNet-18 & MobileNet-V2 & EfficientNet-B0 \\
\midrule
\multirow{3}{*}{ResNet-18}& 
RDED & $42.3 \pm 0.6$ & $40.4 \pm 0.1$ & $36.6 \pm 0.1$ \\
 & D$^3$HR & 
 ${44.3} \pm {0.3}$ & 
 ${42.3} \pm {0.7}$ & 
 ${38.3} \pm {0.2}$ \\
 & 
 \cellcolor{cyan!10} Ours & 
 \cellcolor{cyan!10}$\mathbf{44.6} \pm \mathbf{0.5}$ & 
 \cellcolor{cyan!10}$\mathbf{42.9} \pm \mathbf{0.3}$ & 
 \cellcolor{cyan!10}$\mathbf{38.7} \pm \mathbf{0.1}$ \\
\midrule
\multirow{3}{*}{MobileNet-V2} &
RDED & $34.4 \pm 0.2$ & $33.8 \pm 0.6$ & $28.7 \pm 0.2$ \\
 & D$^3$HR &
 ${43.4} \pm {0.3}$ &
 $\mathbf{46.4} \pm \mathbf{0.2}$ &
 ${37.8} \pm {0.4}$ \\
 & 
 \cellcolor{cyan!10} Ours & 
 \cellcolor{cyan!10}$\mathbf{46.2} \pm \mathbf{0.1}$ & 
 \cellcolor{cyan!10}${45.8} \pm {0.3}$ & 
 \cellcolor{cyan!10}$\mathbf{38.0} \pm \mathbf{0.2}$ \\
\midrule
\multirow{3}{*}{EfficientNet-B0} &
RDED & $22.7 \pm 0.1$ & $21.6 \pm0.2$ & $23.5 \pm 0.3$ \\
& D$^3$HR & ${25.7} \pm {0.4}$ &
${24.8} \pm {0.4}$ &
${28.1} \pm {0.1}$ \\
& 
 \cellcolor{cyan!10} Ours & 
 \cellcolor{cyan!10}$\mathbf{25.9} \pm \mathbf{0.3}$ & 
 \cellcolor{cyan!10}$\mathbf{25.1} \pm \mathbf{0.2}$ & 
 \cellcolor{cyan!10}$\mathbf{28.7} \pm \mathbf{0.4}$ \\
\bottomrule
\end{tabular}
}
\caption{\textbf{Cross-architecture performance on ImageNet-1K, IPC$=$10.}}

\label{tab:cross}
\end{table}
We further evaluate the generalization ability of our method across different network architectures. Table~\ref{tab:cross} reports the Top-1 accuracy on ImageNet-1K (IPC$=$10) when distilled datasets generated by a teacher network are used to train various student architectures, including ResNet-18, MobileNet-V2, and EfficientNet-B0.

Compared to state-of-the-art distillation methods RDED \cite{sun2024diversity} and D$^3$HR \cite{zhao2025taming}, our approach consistently achieves the highest accuracy across almost all of the teacher-student pairs. Notably, when using a MobileNet-V2 teacher, our distilled dataset enables a ResNet-18 student to reach 46.2\%, surpassing both RDED and D$^3$HR by a substantial margin. These results demonstrate that \ours produces realistic representative samples with strong cross-architecture generalization.

\begin{figure*}[t]
    \centering
    \includegraphics[width=0.8\textwidth]{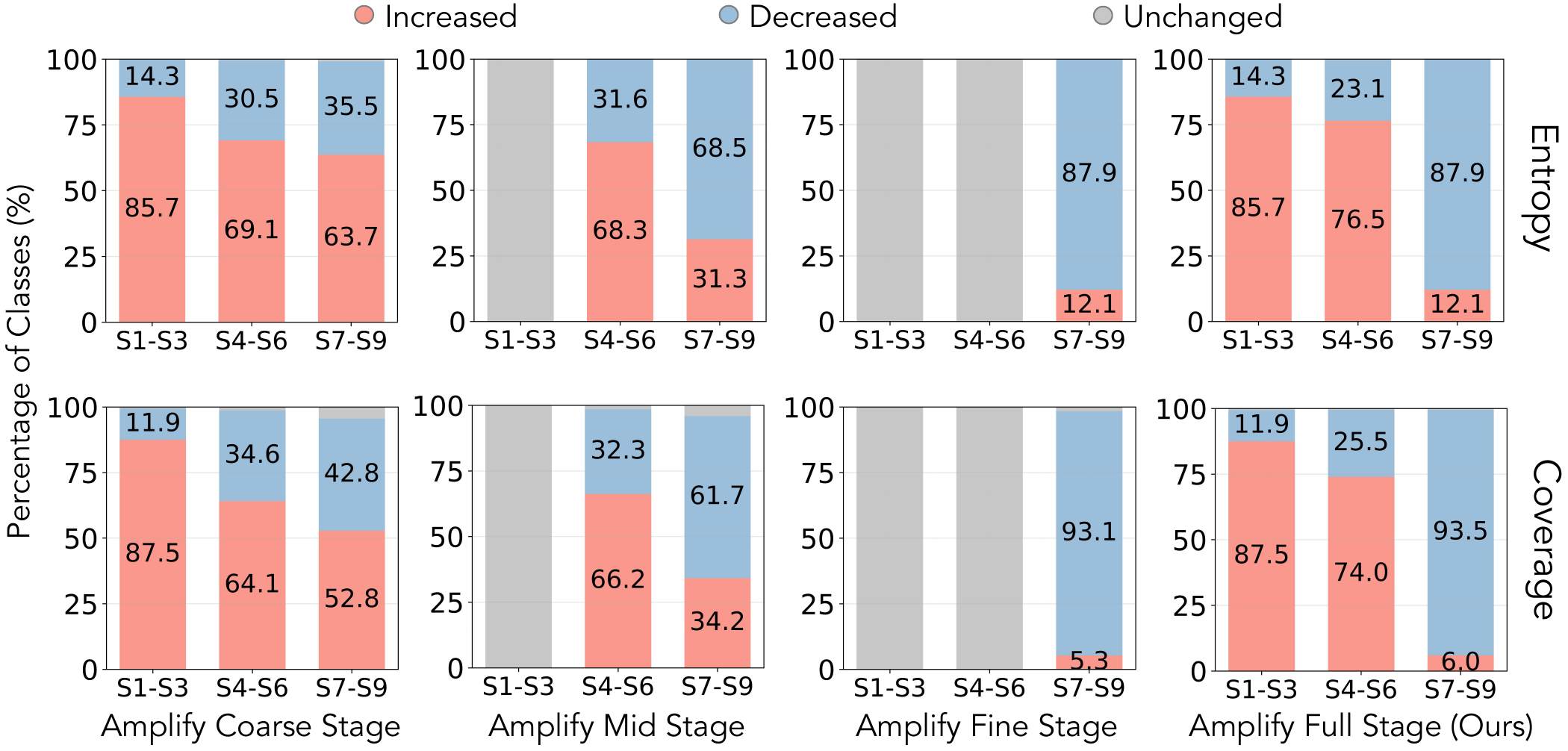}
    \caption{\textbf{Impact of attention amplification strategy on token entropy and coverage on ImageNet-1K, IPC$=$50.}
The histogram shows the percentage of classes whose codebook token entropy and coverage increased, decreased, or remained unchanged after amplifying different stages. Amplifying attention at coarse and mid scales promotes diversity, while fine-scale amplification can concentrate attention.}
    \label{fig:dataset-level-metric}
\end{figure*}

\subsection{Ablation Study}

\noindent \textbf{Effect of Amplification Combination Across Stages.}
\begin{table}[t]
\centering
\resizebox{\columnwidth}{!}{
\begin{tabular}{lcccc}
\toprule
Amp. - $\rho_n\%$ & Coarse & Mid & Fine & Full \\
\midrule
{\color{gray} 0 - 0\%} &  \multicolumn{4}{c}{{\color{gray} $45.6 \pm 0.3$}} \\
\midrule
5 - 30\% & $46.7 \pm 0.2$ & $46.3 \pm 0.1$ & $46.6 \pm 0.3$ & $46.9 \pm 0.2$\\
5 - 70\% & $47.0 \pm 0.3$ & $46.4 \pm 0.2$ & $46.2 \pm 0.3$ & $46.6 \pm 0.2$\\
\midrule
3 - 50\% & $46.8 \pm 0.2$ & $46.7 \pm 0.3$ & $46.5 \pm 0.1$ & $47.1 \pm 0.1$ \\
7 - 50\% & $47.3 \pm 0.2$ & $47.4 \pm 0.3$ & $46.3 \pm 0.2$ & $47.2 \pm 0.3$ \\
\midrule
\cellcolor{cyan!10}5 - 50\% & \cellcolor{cyan!10}$47.6 \pm 0.3$ & \cellcolor{cyan!10}$46.9 \pm 0.1$ & \cellcolor{cyan!10}$46.5 \pm 0.2$ & \cellcolor{cyan!10}$47.6 \pm 0.1$ \\
\bottomrule
\end{tabular}
}
\caption{\textbf{Effect of amplification combination across different stages on ImageNet-1K,
IPC$=$10.} Amp. indicates the amplification number, while $\rho_n\%$ refers to the top $\rho_n\%$ of the $m_N$ regions selected for amplification.}
\label{tab:amplify_strength}
\end{table}

We evaluate the impact of different amplification strategies on ImageNet-1K with IPC$=$10. Specifically, we vary the amplification number (Amp.) and the proportion of top attention regions ($\rho_n\%$) selected for amplification at each stage (Coarse, Mid, Fine, Full). Table~\ref{tab:amplify_strength} reports the accuracy for each configuration.

Without any amplification ($0-0\%$), the baseline performance is $45.6\%$. Introducing selective amplification consistently improves accuracy across all stages. For instance, amplifying 5 regions with 30\% top attention ($5-30\%$) raises the full-stage accuracy to $46.9\%$, while amplifying 3 regions with 50\% top attention ($3-50\%$) further improves it to $47.1\%$. Increasing the amplification number to 5 at 50\% top attention ($5-50\%$) achieves the best overall performance, reaching $47.6\%$ at both the Coarse and Full stages, demonstrating that moderate amplification focused on top attention regions effectively enhances model performance.

These results indicate that carefully combining amplification across hierarchical stages allows the model to better focus on semantically relevant regions, leading to consistent gains over the non-amplified baseline.

\begin{table}[t]
\centering
\resizebox{0.9\columnwidth}{!}{
\begin{tabular}{c|ccc}
\toprule
C-M-F & 0.1-5-5 & 5-0.1-5 & 5-5-0.1 \\
\midrule
 \cellcolor{cyan!10} Top-1 Acc. (\%) &  \cellcolor{cyan!10} $45.9 \pm 0.1$ &  \cellcolor{cyan!10} $46.6 \pm 0.1$ &  \cellcolor{cyan!10} $46.6 \pm 0.3$ \\
\midrule
C-M-F & 0.5-5-5 & 5-0.5-5 & 5-5-0.5 \\
\midrule
 \cellcolor{cyan!10} Top-1 Acc. (\%) &  \cellcolor{cyan!10} $46.3 \pm 0.2$ &  \cellcolor{cyan!10} $46.5 \pm 0.3$ &  \cellcolor{cyan!10}$46.8 \pm 0.3$ \\
\bottomrule
\end{tabular}
}
\caption{\textbf{Ablation on amplification strength across stages on ImageNet-1K,
IPC$=$10.} Each row shows a different setting of Coarse-Mid-Fine (C-M-F) amplification numbers and accuracy.}
\label{tab:amplify_stage_ablation}
\end{table}

\noindent \textbf{Ablation on Amplification Strength Across Stages.}
We investigate the effect of varying amplification strength across hierarchical stages (Coarse, Mid, Fine) on ImageNet-1K with IPC$=$10. Table~\ref{tab:amplify_stage_ablation} reports the Top-1 accuracy for different combinations of amplification numbers applied to the Coarse-Mid-Fine (C-M-F) stages.

From the results, we observe that the model is sensitive to the distribution of amplification across stages. When the Coarse stage receives minimal amplification (0.1) while Mid and Fine stages are heavily amplified (5), the accuracy is relatively low (45.9\%). Increasing the Coarse stage amplification to moderate levels (0.5–5–5) improves performance to 46.3\%, indicating that Coarse-stage attention contributes to capturing global structural information.

Moreover, balancing amplification across stages, e.g., applying higher amplification to Fine stage while maintaining moderate Coarse and Mid stages (5–5–0.5), achieves best accuracy of 46.8\%. These findings suggest that distributing amplification across hierarchical stages allows model to effectively focus on both global and local object features, resulting in improved recognition performance.

\section{Analysis and Insights}
Building on the enhanced empirical results, we further analyze why and how attention amplification improves the distilled data. 
We use the same three-stage split as in the method design. 
Each image can be encoded into a set of discrete codebook token maps across multiple scales using a vector-quantized VAE (VQ-VAE) with a codebook size of 4096~\cite{tian2024visual}. For each token $i$ in the codebook, we can acquire the occurrence count $n_i$ and the normalized occurrence probability $p_i$ of it appearing in a given dataset:
\begin{equation}
p_i=\frac{n_i}{\sum_j n_j}.
\end{equation}
Subsequently, we analyze on the dataset, class, and sample levels to illustrate the influence of attention amplification on different hierarchical stages.

\subsection{Dataset-level}
To study the overall token distribution across classes, we adopt two statistical measures: Entropy~\cite{shannon1948mathematical} and Coverage~\cite{yuan2021bartscoreevaluatinggeneratedtext}. These metrics comprehensively quantify the uncertainty and utilization rate of the codebook token usage, respectively. 
More concretely, we compute these metrics from the occurrence counts of codebook tokens at each scale.
The Entropy $H$ is defined as:
\begin{equation}
H = - \sum_{i=1}^{N} p_i \log p_i.
\label{eq:entropy}
\end{equation}
A higher entropy indicates a more uniform and diverse token distribution.
The Coverage is defined as:
\begin{equation}
\text{Coverage} = \frac{N_{\text{used}}}{N_{\text{total}}},
\label{eq:coverage}
\end{equation}
where $N_{\text{used}}$ is the number of unique tokens that appear at least once, and $N_{\text{total}}$ is 4096, the entire codebook size. 
Higher coverage indicates broader utilization of codebook.


We measure the change in token entropy and coverage across different amplification stages. The percentage of classes with increased, decreased, and unchanged metric values after amplification is summarized in \cref{fig:dataset-level-metric}.
Only amplifying attention in the \textbf{Coarse} stage increases both entropy and coverage for most classes, indicating enhanced token diversity and more distributed token usage. 
Amplifying the \textbf{Mid} stage increases the entropy of entropy and coverage for itself, yet decreases the metric for the Fine stage. Amplification only at the \textbf{Fine} stage leads to more substantial entropy reduction, implying more focused and repetitive token activation. The \textbf{Full} stage amplification (ours) combines all the above effects, increasing entropy for the coarse and mid stages while decreasing that for the fine stage.

These changes indicate that at coarse stages, the semantics of salient regions are rich and diverse. Amplifying those semantics leads to even more different compositions of tokens. Oppositely, at the fine stage, VAR focuses on refining object-specific details. Therefore, after amplification, the token selection is more concentrated. 
Based on the results in \cref{tab:amplify_strength}, the amplification at the coarse stage is more effective than that at the fine stage. This result indicates that coarse-level richness has a more significant influence on the final model performance. 
We include the effects visualization of more amplification settings in the Supplementary Material.

\begin{figure}[t]
    \centering
    \includegraphics[width=\columnwidth]{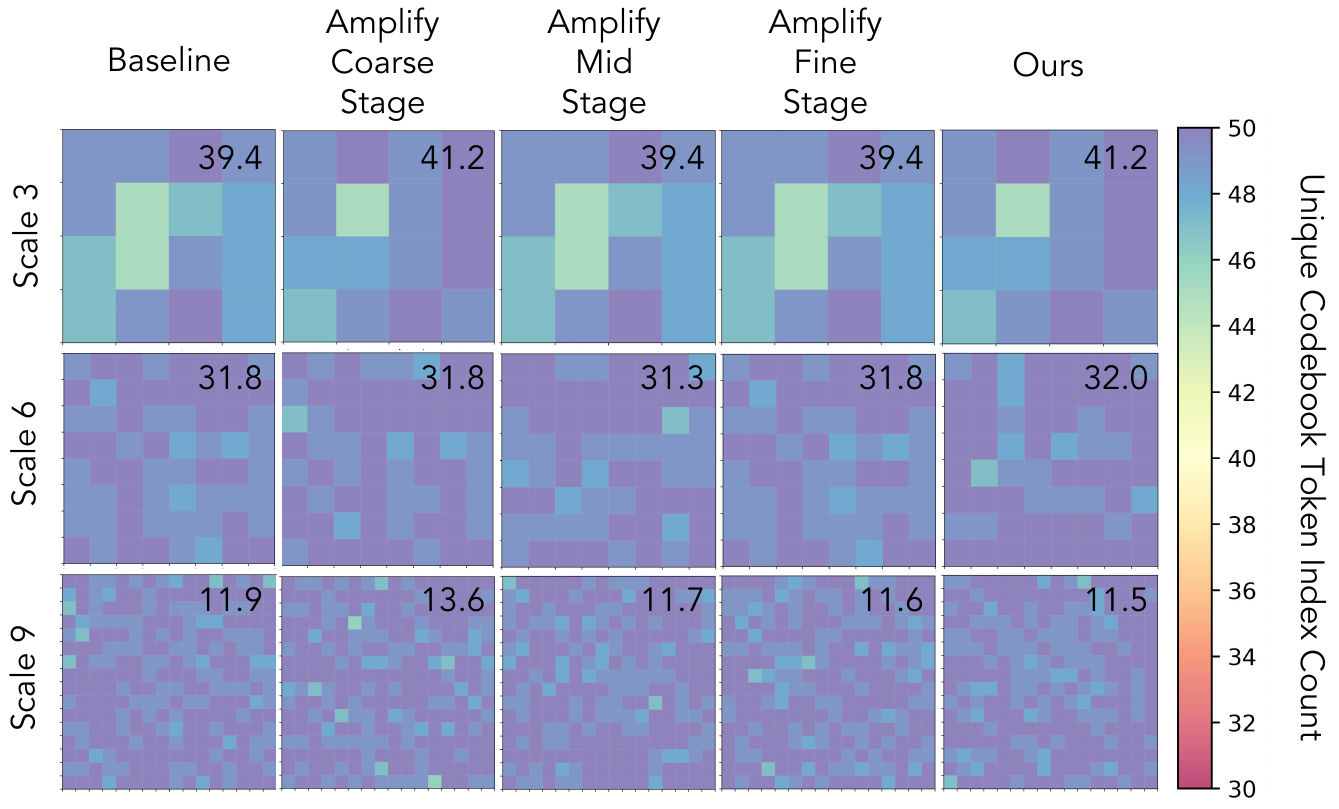}
    \caption{\textbf{Heatmap of unique codebook token occurrences across patch positions and scales on ImageNet-1K, Class 51 – Triceratops,  IPC$=$50}. Darker patches indicate a higher number of unique tokens. The average of unique-token count for each scale is displayed in the upper-right corner of each heatmap.
}
    \label{fig:class}
\end{figure}

\subsection{Class-level}
From \cref{fig:class}, we observe that amplifying coarse stages increases the overall token diversity across patches, evidenced by more dark (e.g. purple) blocks on the heatmap. Amplification on the other two stages only has a moderate influence on per-patch token diversity. 
Our strategy of full amplification aligns with this trend, at coarse and mid stage the model will have more choice of codebook token, which can generate more diverse images; at fine stage, it will refine the details and textures. Although the dataset-level token usage is concentrated at fine scales, within each dataset, the diversity is preserved to maintain rich spatial structures. 

We examine the effect of amplification within a specific class. For \cref{fig:class}, amplifying the \textbf{Coarse} stages increases token diversity across patch positions, reflected by the larger number of dark (e.g., purple) cells in the heatmaps. In contrast, amplification at the \textbf{Mid} and \textbf{Fine} stages induces only moderate changes in per-patch token diversity.

This behavior is consistent with our \textbf{Full} amplification strategy. At the \textbf{Coarse} and \textbf{Mid} stages, amplification expands the set of available codebook tokens, enabling the model to produce more diverse global structures. At the \textbf{Fine} stage, the model instead focuses on refining local details and textures. Although token usage at the dataset level is more concentrated at fine scales, the per-class distributions remain diverse, preserving the spatial richness necessary for high-quality image generation.

\subsection{Sample-level}
We additionally visualize the amplification effects at different stages with specific samples to understand their differences and examine the amplified regions.
Across all the scales, we observe that the attention of the class token is stronger after amplification, indicating that these regions become more class-related. 

\begin{figure}[t]
    \centering
    \includegraphics[width=0.95\columnwidth]{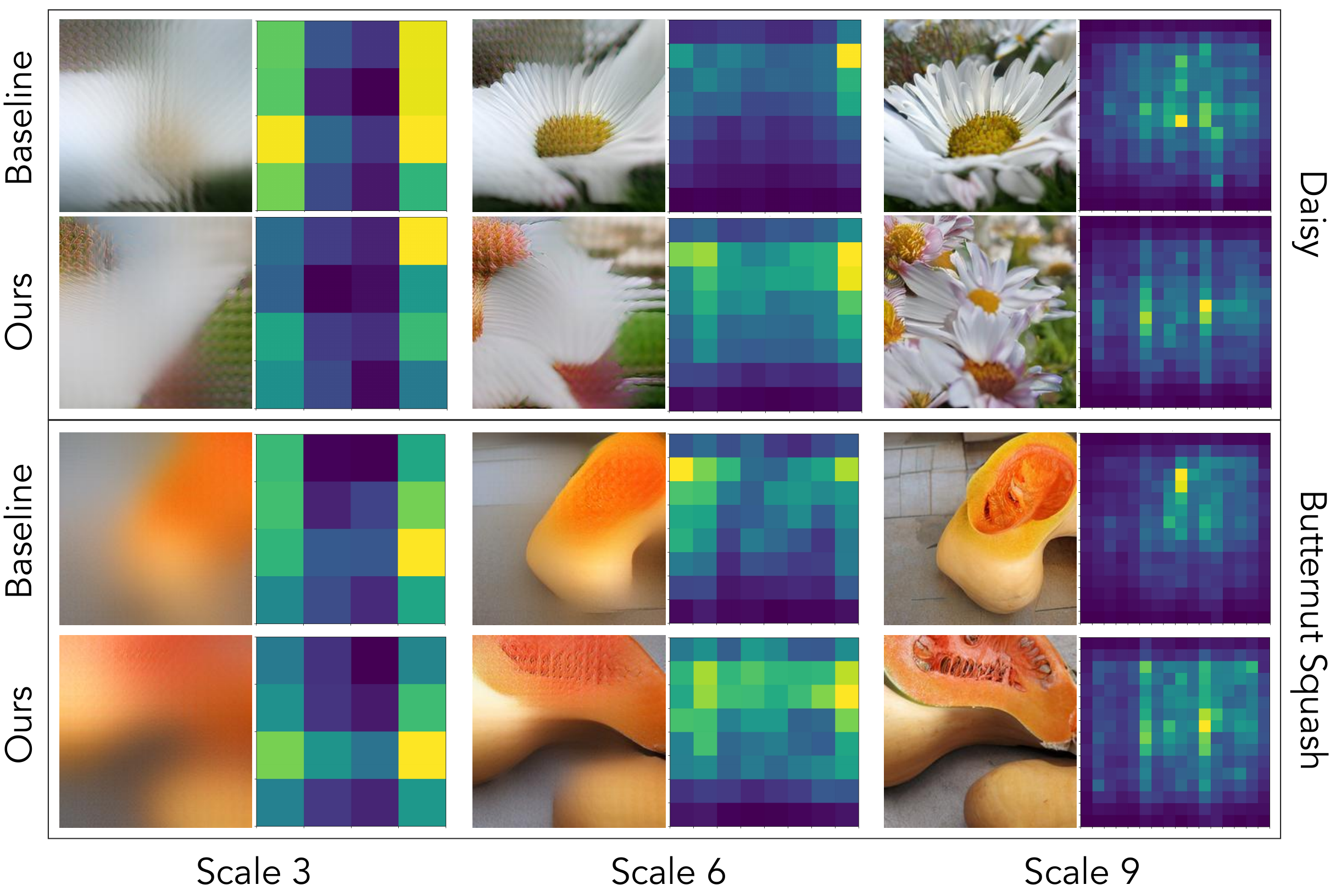}
    \caption{\textbf{Example generated images and attention heatmaps of the class token.} Our method produces richer object details and quantities, achieves stronger semantic alignment, and enhances object-background dependence.
}
    \label{fig:sample}
\end{figure}

\noindent \textbf{Coarse Stage.} At the initial scales, our method primarily captures the overall structure of the object. As illustrated in \cref{fig:sample} at Scale 3, attention is focused on the main object regions, revealing the butternut squash’s distinctive orange hue and the daisy’s white petals. After amplification, the color becomes more vivid, indicating a richer and more diverse selection of codebook tokens. This suggests that the model establishes a strong global representation of the object while maintaining token variability. While these semantics do not directly describe discriminative details, the amplification provides more possible token compositions at later stages. 

\noindent \textbf{Mid Stage.} At intermediate scales, attention becomes more semantically refined. The model selectively emphasizes distinctive object parts, such as the internal and surface features of the squash. \cref{fig:sample} stage 6 shows improved alignment between attention and object semantics, demonstrating that residuals generated at this stage become more closely tied to the target objects. We also observe that combined amplifications at coarse and mid stages can generate multiple object instances, further supporting the view that the amplified semantics are object-related.

\noindent \textbf{Fine Stage.} At fine scales, attention spreads over semantically relevant regions in greater detail. In the squash case, attention is mainly focused on the cross-section of the squash, where finer details of texture and shadows are being added. In comparison, not as much attention is paid to the background or other squashes at the side. It also corresponds to the more concentrated token usage as suggested in \cref{fig:sample}. These refined details provide a moderate performance improvement for training classification models. 

\section{Conclusion}

We propose \ours, a hierarchical attention framework for dataset distillation that leverages the coarse-to-fine structure of VAR models. Our method injects learnable class tokens and amplifies attention at each scale to capture object-level semantics, addressing the lack of semantic guidance in existing distillation approaches. \ours achieves SOTA performance on ImageNet-1K and its subsets, and provides interpretable insights into scale-specific token distributions.

\section*{Acknowledgments}
This research is supported in part by grants from ONR N00014-21-1-2431, NSF OAC-211824, NSF IIS-2310254.

{
    \small
    \bibliographystyle{ieeenat_fullname}
    \bibliography{main}
}

\clearpage
\setcounter{page}{1}
\maketitlesupplementary

\appendix

\section{Algorithm}

We provide more details about our coarse-to-fine autoregressive amplify algorithm here. As shown in \cref{algo: amplify}, we hierarchically amplify the most salient regions at coarse-to-fine scales, yielding semantics that are maximally informative for classification.
We then apply the residual rules described in \cref{sec:pre} to obtain the final output.
In our final configuration, the amplification factor is set to $5$ at all scales and is applied to the weights after the softmax.

\begin{algorithm}[h]
\caption{Coarse-to-Fine Semantic Amplification}
\label{alg:cfa}
\begin{algorithmic}[1]
\State \textbf{Input:} Multi-scale queries $\{Q_n\}_{n=1}^{9}$, keys $\{K_n\}_{n=1}^{9}$, values $\{V_n\}_{n=1}^{9}$; class token is appended as the last query in $Q_n$, Per-head mask $\{p_n^{(h)}\}$; per-scale key counts $\{L_k^n\}$; heads $H$, head dim $d_h$, Stage schedules $(\rho_{1{:}3},\rho_{4{:}6},\rho_{7{:}9})$ and $(\beta_{1{:}3},\beta_{4{:}6},\beta_{7{:}9})$.
\State \textbf{Output:} Amplified attentions $\{\tilde{\alpha}_n\}$ and/or reweighted contexts $\{\tilde{O}_n\}$.
\For{$n = 1$ to $9$} \\
\Comment{coarse (1–3) $\rightarrow$ mid (4–6) $\rightarrow$ fine (7–9)}
    \State $(\rho,\beta) \gets \textsc{StageParams}(n)$
    \For{$h = 1$ to $H$}
        \State $L_n^{(h)} \gets \dfrac{Q_n^{(h)} (K_n^{(h)})^\top}{\sqrt{d_h}} + p_n^{(h)}$ \\
        \Comment{masked logits at scale $n$}
        \State $\alpha^{(h)}_{n,\mathrm{cls}} \gets \mathrm{Softmax}\!\big(L_n^{(h)}[-1,\,1{:}L_k^n]\big)$ \\
        \Comment{class $\rightarrow$ same-scale keys}
    \EndFor
    \State $m_n \gets \frac{1}{H}\sum_{h=1}^{H}\alpha^{(h)}_{n,\mathrm{cls}} \in \mathbb{R}^{1\times L_k^n}$ \\
    \Comment{head-avg saliency}
    \State $k \gets \max\!\big(1,\lfloor \rho \cdot L_k^n \rfloor\big)$
    \State $S_n \gets \textsc{TopK}(m_n, k)$
    \State $a_n \in \{0,1\}^{L_k^n}$ initialized to $0$
    \State $(a_n)_j \gets 1$ iff $j \in S_n$ \\
    \Comment{binary indicator}
    \For{$h = 1$ to $H$}
        \State $B_n^{(h)} \gets \beta \cdot \mathbf{1}_{L_q^n+1}\, a_n^\top \in \mathbb{R}^{(L_q^n+1)\times L_k^n}$ \\
        \State $\tilde{L}_n^{(h)} \gets L_n^{(h)}$
        \State $\tilde{L}_n^{(h)}[:,\,1{:}L_k^n] \mathrel{+}= B_n^{(h)}$
        \State $\tilde{\alpha}_n^{(h)} \gets \mathrm{Softmax}(\tilde{L}_n^{(h)})$
    \EndFor
    \State $\tilde{O}_n \gets \textsc{Attnout}\!\big(\{\tilde{\alpha}_n^{(h)}\}_{h=1}^{H},\,\{V_n^{(h)}\}_{h=1}^{H}\big)$ \\
\EndFor
\State \Return  $\{\tilde{O}_n\}_{n=1}^{9}$
\label{algo: amplify}
\end{algorithmic}
\end{algorithm}





\section{Generalizing to DiT}
\label{sec:DiT}
To demonstrate that HIERAMP is backbone-agnostic, we extended it to a Diffusion Transformer (DiT)~\cite{peebles2023scalable} and evaluated it on ImageNet-Woof~\cite{Howard_Imagewoof_2019}.
Specifically, let $A$ denote the attention output at a given transformer block. 
We apply spatially guided scaling:
\begin{equation}
\tilde{A} = A \odot (1 + \alpha M),
\end{equation}
where $M$ is the object-region projected to the token space, $\alpha$ controls the amplification strength, and $\odot$ denotes element-wise multiplication. As shown in 
Table~\ref{tab:dit}, \ours improves the accuracy while maintaining stable generation compared to the vanilla DiT baseline, which indicates that our method generalizes beyond VAR and is compatible with diffusion-based transformer backbones.

\begin{figure}[t]
    \centering
    \includegraphics[width=0.8\columnwidth]{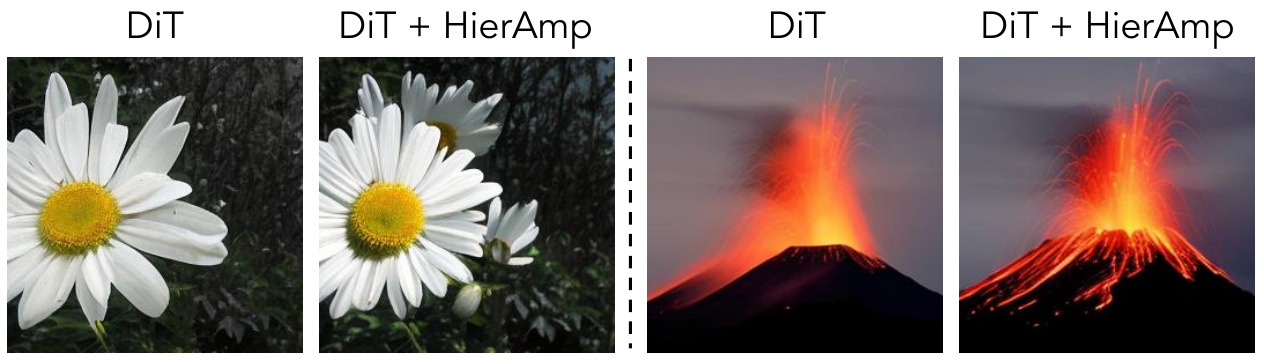}
    \caption{\textbf{Qualitative comparison of the DiT backbone before and after applying HIERAMP.}
    }
    \label{fig:dit}
\end{figure}

Figure~\ref{fig:dit} shows visual comparisons before and after applying \ours. 
We observe enhanced object prominence and clearer semantic structures, while background regions remain largely unaffected or much more relevant. 
The modulation improves object-level consistency without introducing noticeable artifacts, demonstrating the effectiveness of region-aware scaling in diffusion transformers.

Importantly, this extension requires no architectural redesign and introduces negligible computational overhead. 
These findings suggest that \ours provides a general mechanism for hierarchical semantic control across diverse generative backbones.

\section{FID Comparisons}
\label{sec:fid}
\paragraph{Comparison with Prior Methods}
We compare the Fréchet Inception Distance (FID)~\cite{heusel2018ganstrainedtimescaleupdate} of our method with representative dataset distillation approaches, including Minimax~\cite{gu2024efficient} and D$^3$HR~\cite{zhao2025taming}. 
FID is computed against the original ImageNet-1K training set under 10 and 50 images per class (IPC). Lower FID indicates better performance.

As shown in Table~\ref{tab:fid}, our method consistently achieves lower FID scores across both IPC settings. 
In particular, at 10 IPC, our approach improves FID from 18.3 (Minimax) and 19.0 (D$^3$HR) to 17.3. 
At 50 IPC, we obtain 13.2, outperforming others. We further analyze whether the hierarchical amplification strategy affects generative fidelity. 
Table~\ref{tab:fid} reports FID before and after applying \ours on VAR.
The results show that FID remains comparable to the default VAR baseline across IPC settings. 
Specifically, the difference is marginal (e.g., 17.5 vs. 17.3 at 10 IPC).

The above results indicate that \ours preserves visual fidelity while enhancing semantic discriminability. 
This confirms that the proposed strategy does not degrade generation quality.

\begin{table}[t]
\centering
\setlength{\tabcolsep}{4pt} 
\renewcommand{\arraystretch}{0.9} 
\scriptsize 
\begin{tabular}{lcc|cc}
\toprule
Dataset & Model & IPC & DiT & DiT + \ours \\
\midrule
\multirow{2}{*}{ImageNet-Woof} &\multirow{2}{*}{ResNet-18}& 10 & $41.0 \pm 0.6$ & $\mathbf{43.1}\pm \mathbf{0.3}$ \\
& & 50 & $ 66.2\pm 0.3$ & $\mathbf{68.2}\pm \mathbf{0.3}$ \\
\bottomrule
\end{tabular}
\caption{\textbf{Quantitative comparison of the DiT backbone before and after applying HIERAMP.}}
\label{tab:dit}
\end{table}

\begin{table}[t]
\centering
\vspace{-3mm}
\setlength{\tabcolsep}{3pt} 
\renewcommand{\arraystretch}{0.85} 
\scriptsize 
\begin{tabular}{l|cccc}
\toprule
IPC & Minimax & D$^3$HR & VAR & Ours \\
\midrule
10 & $18.3 \pm 0.2$ & $19.0 \pm 0.2$ & $17.5\pm0.1$& $\mathbf{17.3 \pm 0.1}$ \\
50 & $14.3 \pm 0.2$ & $14.9 \pm 0.1$ & $\mathbf{13.1\pm0.1}$ &$13.2 \pm 0.1$\\
\bottomrule
\end{tabular}
\caption{\textbf{FID of different dataset distillation methods on ImageNet-1K under 10 and 50 IPC.}}
\label{tab:fid}
\end{table}

\section{Effectiveness of Class Tokens}
We evaluate the impact of class tokens on both performance and generation quality of VAR. As shown in \cref{tab:cls_token_ablation}, models trained with and without class tokens exhibit highly similar top-1 accuracy across different IPC settings, indicating that class tokens introduce no significant advantage or degradation in classification performance.

\begin{figure*}[t]
    \centering
\includegraphics[width=\textwidth]{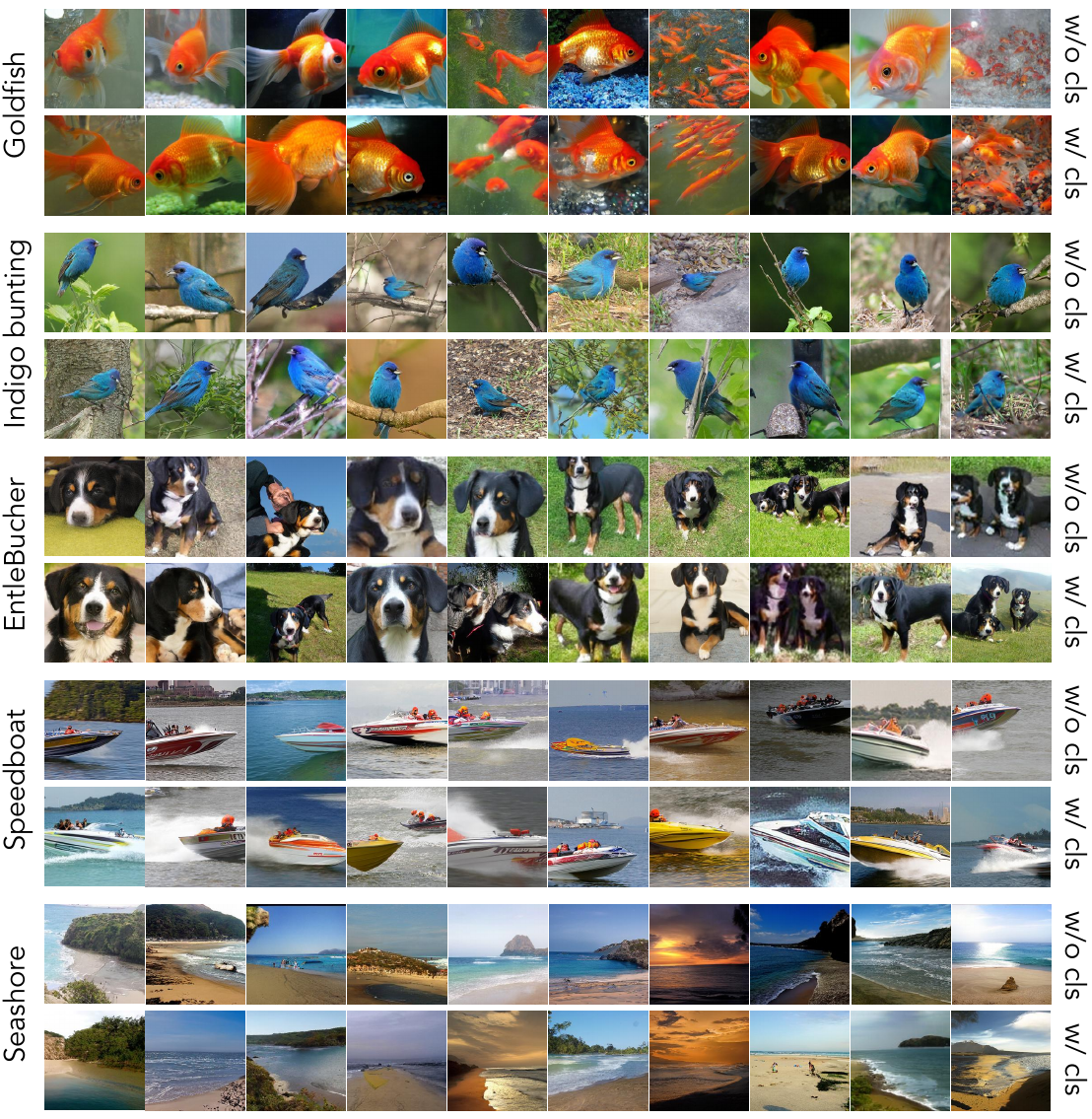}
    \caption{\textbf{Visualization of images generated by VAR with and without class tokens on ImageNet-1K, IPC$=$10.} “w/o” denotes VAR without class tokens, and “w/” denotes VAR with class tokens. The results indicate comparable generative capacity between the two models.
}
    \label{fig:with_without_cls}
\end{figure*}

To further examine their generative behavior, we visualize distilled images produced by both variants in \cref{fig:with_without_cls}. The “w/o” setting denotes VAR trained without class tokens, while “w/” denotes the standard model with class tokens. Consistent with the quantitative results, the visualizations demonstrate comparable generative capacity: both models produce class-consistent images with similar semantic fidelity and structural detail. These findings show that employing VAR with class tokens is a reliable design choice, and can future provide additional object-focused attention benefits.

\begin{table}[h]
\centering
\footnotesize
\renewcommand{\arraystretch}{1.0}
\begin{tabular}{c|cc}
\toprule
IPC & VAR without class tokens & VAR with class tokens \\
\midrule
10 & $45.9\pm0.3$ & $45.6\pm0.3$ \\
50 & $59.5\pm0.1$ & $59.3\pm0.1$ \\ 
\bottomrule
\end{tabular}
\caption{\textbf{Effect of class tokens on top-1 accuracy on ImageNet-1K.} Models with and without class tokens show comparable performance across different IPC settings.}
\label{tab:cls_token_ablation}
\end{table}

\section{Computational latency}
\label{sec:latency}
\paragraph{Comparison with Diffusion Models}
We compare the inference speed of our method with a representative diffusion model, DDIM~\cite{songdenoising} used in D$^3$HR, under the same setting (batch size = 1).
As shown in Table~\ref{tab:vs}, our approach achieves significantly lower latency, processing an image in $0.147$ s compared to $0.456$ s for DDIM with 30 denoising steps. 
This efficiency arises from the progressive prediction of scales using fewer tokens and a reduced number of inference steps ($\leq 10$), demonstrating that hierarchical amplification can accelerate generation without sacrificing quality.

\begin{table}[t]
\centering
\vspace{-3mm}
\setlength{\tabcolsep}{4pt}
\renewcommand{\arraystretch}{0.9}
\scriptsize
\begin{tabular}{l|cc}
\toprule
Model  & Ours & D$^3$HR (DDIM-based, 30 steps) \\
\midrule
Time (s/img) & $ \mathbf{0.147}\pm \mathbf{0.001} $  & $0.456\pm0.002$  \\
\bottomrule
\end{tabular}
\caption{\textbf{Inference latency comparison with DDIM-based method on ImageNet-Woof.}}
\label{tab:vs}
\end{table}

\paragraph{Resource Consumption of Distillation}
We further report the computational cost of our dataset distillation pipeline based on VAR. 
Table~\ref{tab:time} shows latency and peak memory for the base VAR, VAR with the class token, and VAR with class token plus hierarchical attention amplification (\ours). 
The additional overhead introduced by the class token and attention modulation is negligible, with runtime increasing only slightly (0.139 $\rightarrow$ 0.147 s/img) and peak memory remaining comparable (1.770 $\rightarrow$ 1.840 GB).

\begin{table}[t]
\centering
\setlength{\tabcolsep}{4pt}
\renewcommand{\arraystretch}{0.9}
\footnotesize
\resizebox{\linewidth}{!}{
\begin{tabular}{l|ccc}
\toprule
Model  & Base VAR & VAR + cls & VAR + cls + Amp (Ours)\\
\midrule
Latency (s/img) & $0.139\pm0.002 $ & $0.145\pm0.002$ & $ 0.147\pm 0.001$   \\
Peak Memory (GB) & $1.770\pm0.000 $ & $1.790\pm0.000$ & $ 1.840\pm0.000 $   \\
\bottomrule
\end{tabular}
}
\caption{\textbf{Latency and peak memory for VAR-based distillation with incremental modules. }}
\label{tab:time}
\vspace{-3mm}
\end{table}

These results indicate that \ours provides a computationally efficient alternative to standard diffusion-based generation while maintaining competitive fidelity. 
The progressive scale prediction and token-efficient design contribute to faster inference, and the pipeline ensures minimal runtime and memory overhead for extended VAR variants.

\section{Analysis on More Amplification Combinations}

\begin{figure*}[t!]
    \centering
\includegraphics[width=0.9\textwidth]{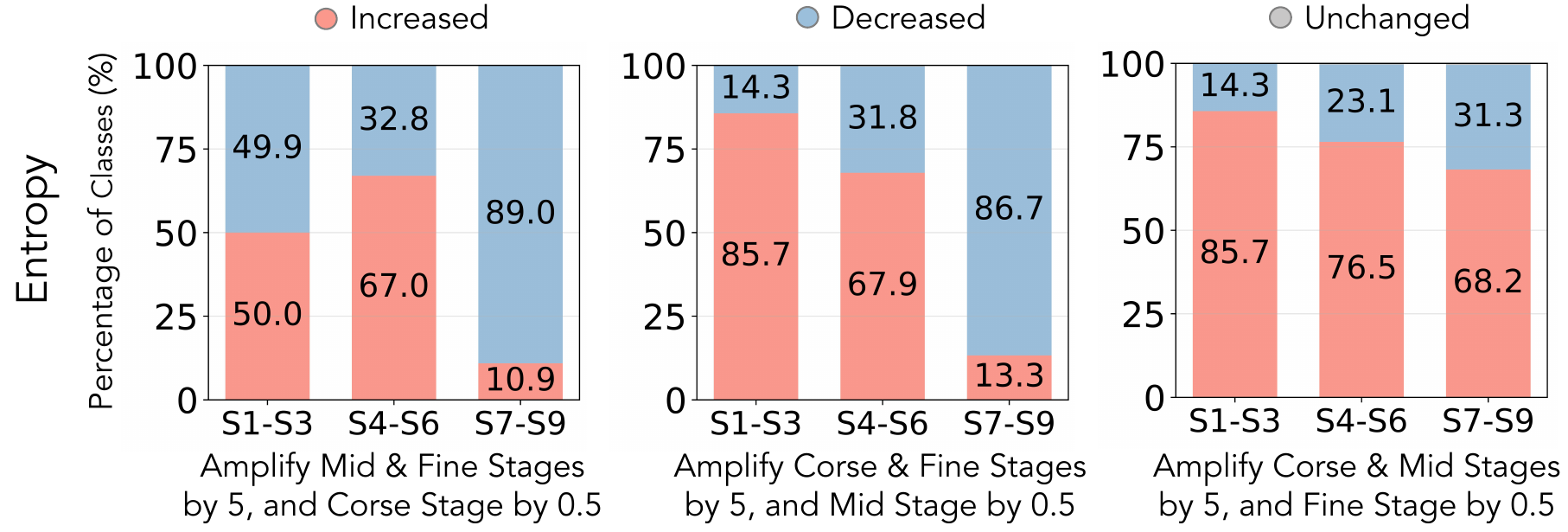}
    \caption{\textbf{More amplification combinations impact of attention amplification strategy on token entropy on ImageNet-1K, IPC$=$50.}
}
    \label{fig:entropy_appendix}
\end{figure*}

\begin{figure*}[h!]
    \centering
\includegraphics[width=0.9\textwidth]{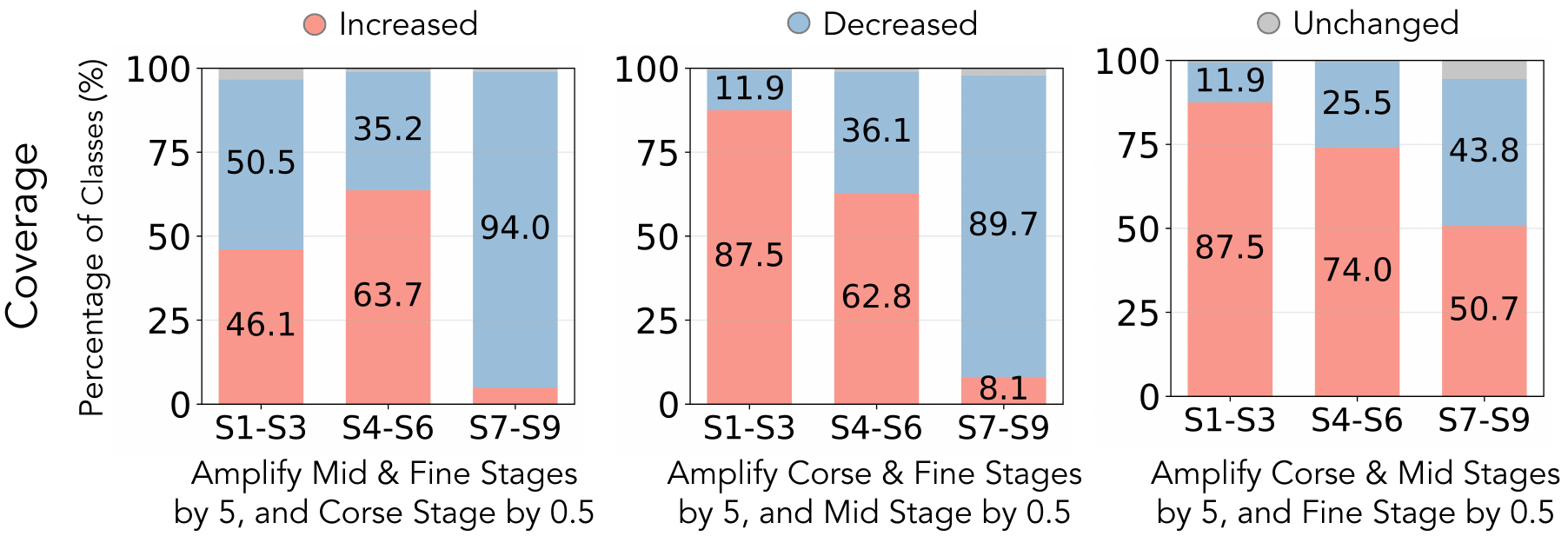}
    \caption{\textbf{More amplification combinations impact of attention amplification strategy on token coverage on ImageNet-1K, IPC$=$50.}
}
    \label{fig:coverage_appendix}
\end{figure*}

We conducted additional experiments with different attention amplification combinations to further validate the conclusions we draw from \cref{fig:dataset-level-metric}: (1) amplify Mid \& Fine stages by 5 and Coarse stage by 0.5; (2) amplify Coarse \& Fine stages by 5 and Mid stage by 0.5; (3) amplify Coarse \& Mid stages by 5 and Fine stage by 0.5. As shown in \cref{fig:dataset-level-metric}, amplifying attention at coarse and mid scales increases diversity, with many classes exhibiting higher entropy, whereas fine-scale amplification concentrates attention, with many classes exhibiting lower entropy.

For combination (1), which is similar to full-stage amplification (S1–S9) except that the Coarse stage is not amplified by 5, but 0.5, we expect fewer classes to exhibit increased token entropy in S1–S3, consistent with the results in \cref{fig:entropy_appendix}. Similarly, in combination (2), fewer classes show increased token entropy in S4–S6 compared to the full-stage case in \cref{fig:dataset-level-metric}. Additionally, reducing the amplification factor in the Fine stage leads to a larger percentage of classes with increased token entropy compared to S7–S9 in \cref{fig:dataset-level-metric}, indicating that smaller amplification at fine scales results in less concentrated attention.

The coverage results in \cref{fig:coverage_appendix} further corroborate these trends. When amplification is applied to the Coarse and Mid stages, coverage expands across a larger portion of the codebook tokens, reflecting greater diversity in the attended regions. In contrast, stronger amplification at the Fine stage leads to more focused and localized attention, reducing overall coverage. For combination (1), the reduced amplification at the Coarse stage results in lower coverage gains in S1–S3 compared to full-stage amplification. Similarly, combination (2) yields smaller coverage increases in S4–S6, aligning with the patterns observed in the entropy analysis. Finally, combination (3), which applies a weaker amplification to the Fine stage, produces broader coverage in S7–S9 relative to the full-stage setting, consistent with the observation that smaller fine-scale amplification reduces attention concentration.

\section{Image Visualization and Comparison}

\begin{figure*}[h]
    \centering
\includegraphics[width=\linewidth]{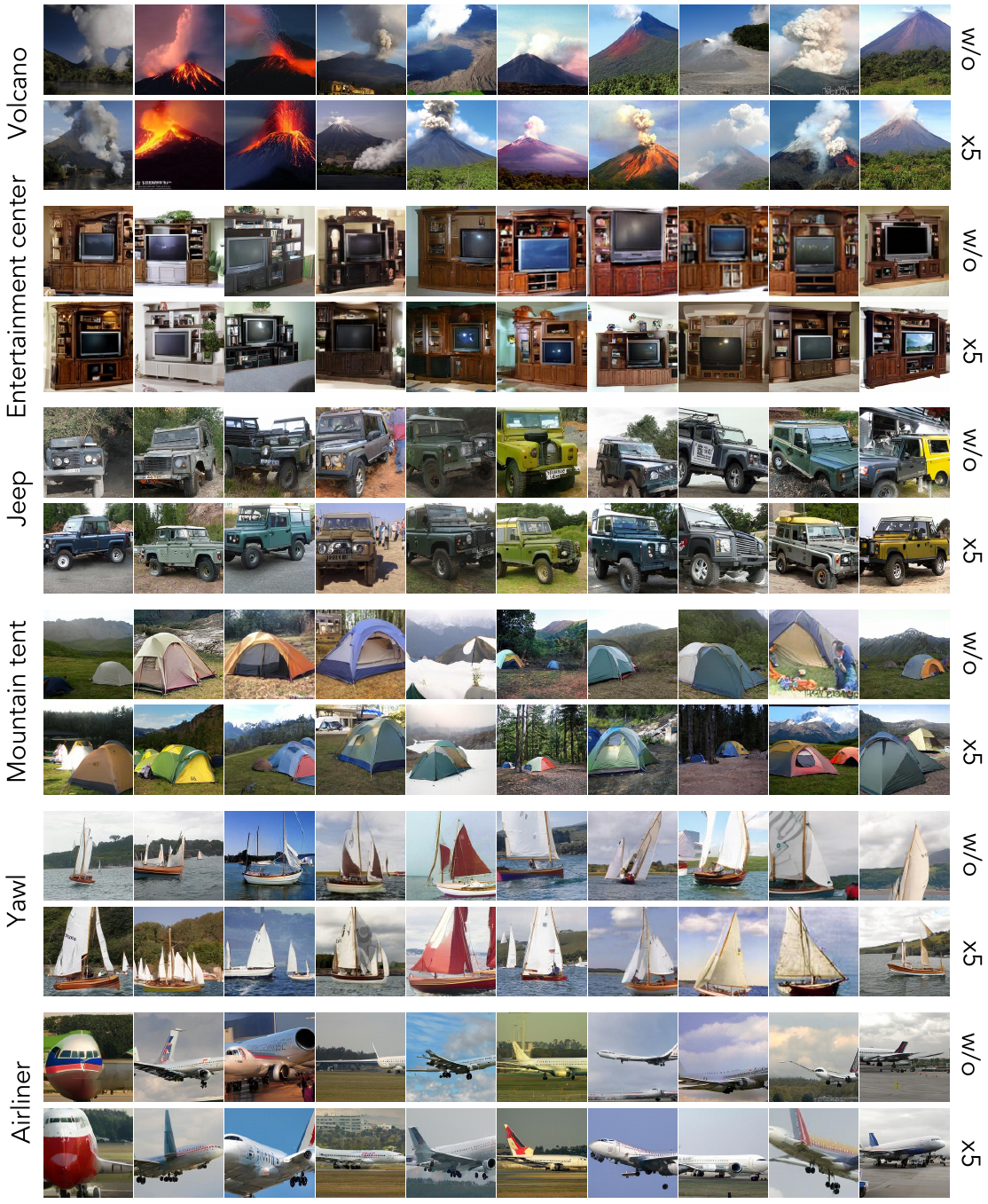}
    \caption{\textbf{Visualization of the generated distilled images ($224\times224$) on ImageNet-1K, IPC$=$10.} The first row shows distilled images without amplification (VAR with class tokens). The second row shows distilled images with amplification applied to Full stages (1-9) by a amplification factor of 5.
}
    \label{fig:vis1}
\end{figure*}

\begin{figure*}[h]
    \centering
\includegraphics[width=\linewidth]{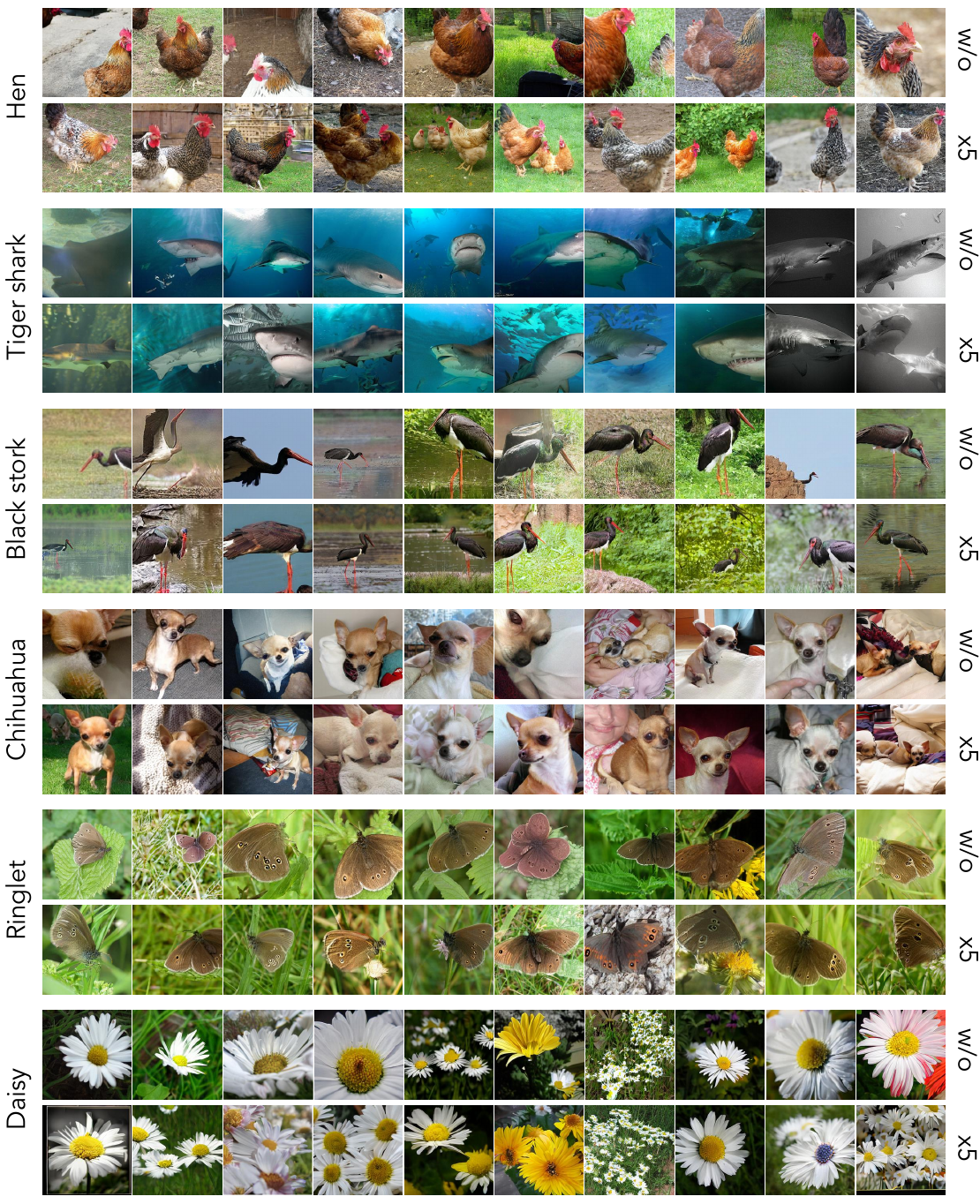}
    \caption{\textbf{Visualization of the generated distilled images ($224\times224$) on ImageNet-1K, IPC$=$10.} The first row shows distilled images without amplification (VAR with class tokens). The second row shows distilled images with amplification applied to Full stages (1-9) by a amplification factor of 5.
}
    \label{fig:vis2}
\end{figure*}

We show further visualizations of the distilled images in this section. As illustrated in \cref{fig:vis1} and \cref{fig:vis2}, \ours generates finer object detail and more diverse objects in a single image, better semantic alignment, and stronger object–background coupling for each class, providing an effective representation of the full dataset.

\end{document}